\documentclass[10pt,twocolumn,letterpaper]{article}

\usepackage{iccv}
\usepackage{times}
\usepackage{epsfig}
\usepackage{graphicx}
\usepackage{amsmath}
\usepackage{amssymb}
\usepackage{subfig}
\usepackage{pdfpages}

\usepackage[pagebackref=true,breaklinks=true,letterpaper=true,colorlinks,bookmarks=false]{hyperref}

 \iccvfinalcopy 

\def\iccvPaperID{629} 

\ificcvfinal\fi
\begin{document}

\title{MegaFace: A Million Faces for Recognition at Scale}
 
\author{D. Miller ~~~~  E. Brossard~~~~ S. Seitz ~~~~ I. Kemelmacher-Shlizerman \\
Dept. of Computer Science and Engineering\\
University of Washington}

	\teaser{
	
		\includegraphics[width=.9\linewidth]{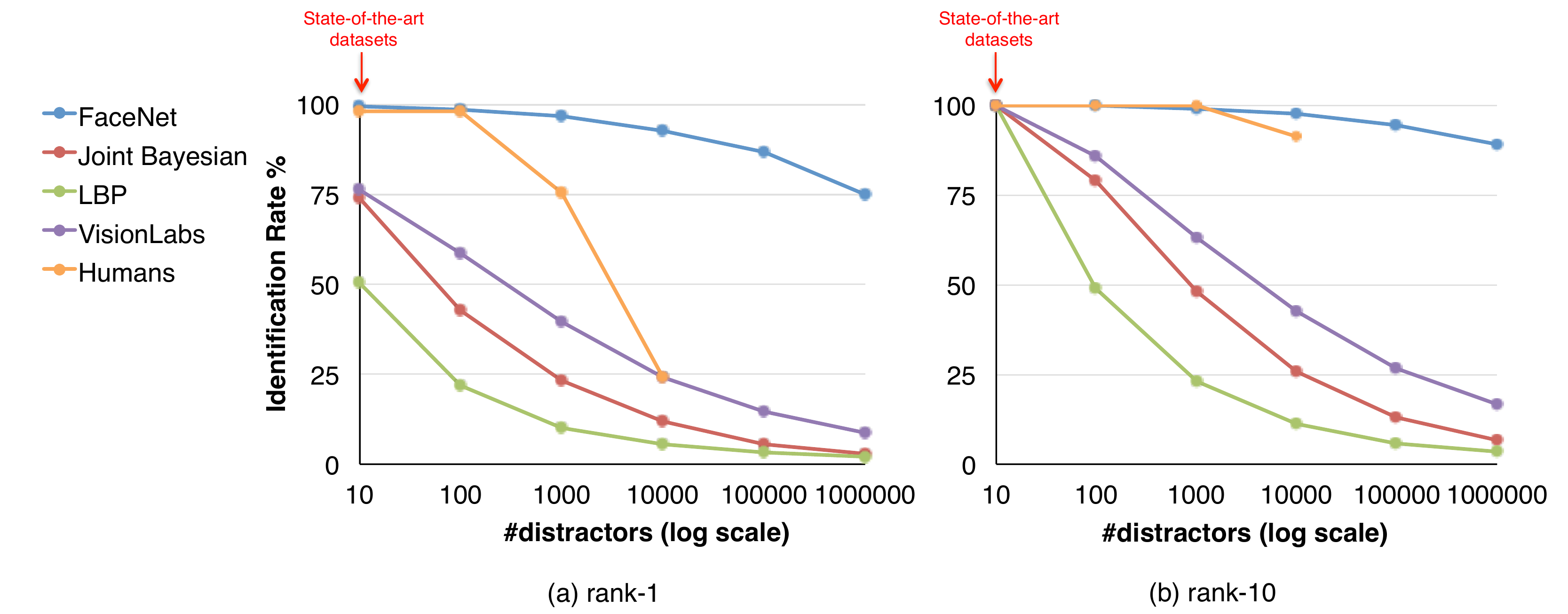}
		\caption{We evaluate how recognition performs with increasing numbers of faces in the database: (a) shows rank-1 identification rates, and (b) rank-10.  Recognition rates drop once the number of distractors increases. We also present first large-scale human recognition results (up to 10K distractors). Interestingly,    Google's deep learning based FaceNet is more robust at scale than humans. See \small{\url{http://megaface.cs.washington.edu}} to participate in the challenge.}
		\label{fig:teaser}

	}
	
\maketitle

\begin{abstract}
		Recent face recognition experiments on the LFW~\cite{huang2007labeled} benchmark show that face recognition is performing stunningly well, surpassing human recognition rates. In this paper, we study face recognition at scale. Specifically, we have collected from Flickr a \textbf{Million} faces and evaluated state of the art face recognition algorithms on this dataset. We found that the performance of algorithms varies--while all perform great on LFW, once evaluated at scale recognition rates drop drastically for most algorithms. Interestingly, deep learning based approach by \cite{schroff2015facenet} performs much better, but still gets less robust at scale. We consider both verification and identification problems,  and evaluate how pose  affects recognition at scale. Moreover, we ran an extensive human study on Mechanical Turk to evaluate human recognition at scale, and report results. All the photos are creative commons photos and are released for research and further experiments on \small{\url{http://megaface.cs.washington.edu}}.
\end{abstract}


\section{Introduction}


Face recognition has seen major breakthroughs in the last couple of years, with new results by multiple groups   \cite{schroff2015facenet,taigman2014deepface,sun2015deepid3} {\bf surpassing human performance} on the leading 
Labeled Faces in the Wild (LFW) benchmark \cite{huang2007labeled} and achieving near perfect results.  

Is face recognition solved?  
Many applications require accurate identification at {\em planetary scale}, i.e., finding the best matching face in a database of billions of people.  This is truly like finding a needle in a haystack.  Face recognition algorithms did not deliver when the police was searching for the suspect of the Boston marathon bombing~\cite{klontz2013case}.  Similarly, do you believe that current cell-phone face unlocking programs will protect you against anyone on the planet who might find your lost phone?  These and other face recognition applications require finding the true positive match(es) with negligible false positives.

In this paper, we introduce the {\em Megaface} dataset and benchmark for large scale face recognition.  
The goal of this dataset is to evaluate the performance of current face recognition algorithms with up to a million {\em distractors}, i.e., up to a million people who are not in the test set.  Our key objectives for this dataset are that it should 1) contain photos ``in the wild'', i.e., with unconstrained pose, expression, lighting, and exposure, 2) contain regular people, not easily recognizable celebrities, 3) be broad rather than deep, i.e., contain many different people rather than many photos of a small number of people, and 4) be publicly available, to enable distribution within the research community.  Whereas previous face benchmarks have relied on celebrity photos, passport photos, or mugshots, our objectives require a different approach.  Instead, we leverage the recently released database of Flickr Creative Commons photos \cite{thomee2015new}, from which we extracted 1 million faces (randomly sampling the full 100M photo collection).  We intend to release even larger datasets (from the full 100M collection) in the future, setting aside training and testing sets to ensure an even playing field. 

Based on this new benchmark, we address fundamental questions and introduce the following key findings:
\begin{itemize}
\item {\bf How well do current face recognition algorithms scale?}  
{\bf Key finding:}  
While performance {\bf drops by 70\%} for most algorithms, Google's~\cite{schroff2015facenet} deep-learning based FaceNet achieves 
75\% identification rate even with a million distractors (Fig.~\ref{fig:teaser}).
\item {\bf How well does human face recognition scale?}  Even devising a practical human face identification experiment (requiring people to sort lists containing thousands of faces) is challenging.
We performed the first large scale human face identification experiment, leveraging a crowd of Mechanical Workers to collectively sort the best matches from each probe image against a database containing one true match and ten thousand distractors. Humans' rank-1 identification rate is 23.9\% with 10K distractors,  91.13\% at rank-10. 

\item {\bf How does pose affect recognition performance?}  Somewhat surprisingly, recognition rates {\bf drop} when comparing frontal-to-frontal images, compared to the task of comparing faces when the pose is not controlled.
\end{itemize}

While this paper benchmarks only a sparse sampling of face recognition algorithms, we believe that it's a reasonable initial sampling, as it includes the current top performer on LFW \cite{schroff2015facenet}, another commercial LFW top-performer (VisionLabs), and two baseline algorithms (Joint Bayesian and LBP) that are popular in the academic community.
Furthermore, if the paper is accepted to ICCV, we will maintain and update this benchmark online and solicit contributions from the other top performers (e.g., we are in contact with Facebook and hope to include DeepFace \cite{taigman2014deepface} and others before the paper goes out to press).

The remainder of this paper is organized as follows.  We first describe related face datasets and benchmarking efforts, and the details of our dataset.  We then describe our evaluation methodology, our efforts at evaluating human performance at large scale, and our benchmark of face recognition algorithms.  We then analyze the effects of pose and recognition accuracy, and conclude the paper.

\section{Related Work}

Early work in face recognition focused on controlled datasets where subset of lighting, pose, or facial expression was kept fixed, e.g.,  \cite{georghiades1997yale, gross2010multi}. With the advance of algorithms, the focus moved to unconstrained scenarios with a number of important benchmarks appearing,  e.g., FRGC, Caltech Faces, and many more (see \cite{huang2007labeled} Fig 3. for a list of all the databases), as well as,  thorough evaluations were made \cite{grother2010report,zhao2003face}.  A big challenge, however,  was to collect photos of large number of individuals. 

In 2007, Huang et al. \cite{huang2007labeled} created the benchmark Labeled Faces in the Wild (LFW). The LFW database includes 13K photos of 5K different people. It was collected by running Viola-Jones face detection \cite{viola2004robust} on  Yahoo News photos. LFW captures  celebrities photographed under completely unconstrained conditions (arbitrary lighting,  pose, and expression) and it  turned out to be an amazing resource for face analysis community. Since 2007, a number of databases appeared that include larger numbers of photos per person (LFW has 1620 people with more than 2 photos), video information, and even 3D information \cite{kumar2009attribute, beveridge2013challenge, yi2014learning, wolf2011face, chen2012bayesian,  ng265data}.   However, LFW is still the leading benchmark on which all state of the art recognition methods are evaluated and compared. Indeed, just in the last few months a number of methods \cite{schroff2015facenet,sun2014deeply,sun2015deepid3,taigman2014deepface,taigman2014web} reported recognition rates above 99\%+ \cite{hu2015face} (better than human recognition rates estimated on the same dataset by \cite{kumar2011describable}).The perfect recognition rate on LFW is 99.9\% (it is not 100\% since there are 5 pairs of photos that are mislabeled).

\begin{figure}
	\begin{center}
		\includegraphics[width=1\linewidth]{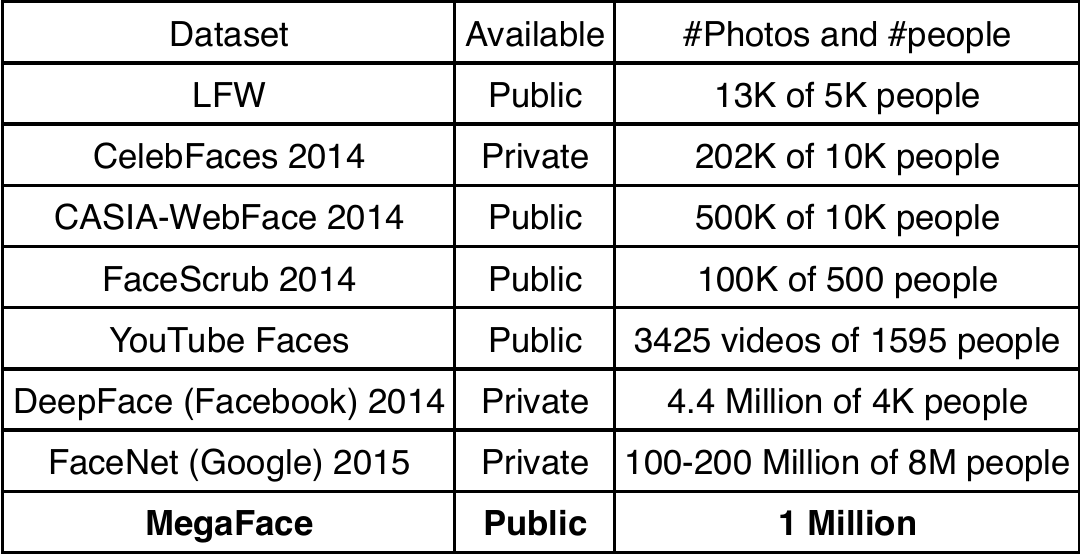}
	\end{center}
	\caption{Representative sample of face recognition datasets that were created in the recent  years (in addition to LFW). All the public datasets are small scale, and all the large scale datasets are mainly used for training rather than testing and are not publicly available. MegaFace (this paper) is the first large scale unconstrained dataset. It is collected from Flickr and will be available publicly.}
	\label{fig:datasets}
\end{figure}

Our paper is about taking the face recognition task to the next level and revealing the challenges that appear once recognition is done on large scale (many orders of magnitude larger than current evaluation).  Specifically, when the number of photos in the database includes 1 Million faces versus tens of thousands (Fig~\ref{fig:datasets}). Companies like Facebook and Google have access to extremely large numbers of photos, e.g., Facebook has trained on  4 Million photos of 4K people \cite{taigman2014deepface}, Google \cite{schroff2015facenet}
trained on 200 Million photos of 8 Million people. These datasets, however, are not available to the public and were used only for training and not testing. Large scale evaluations were performed only on controlled datasets (visa photographs, mugshots, lab captured photos) by NIST \cite{grother2010report}, and report recognition results of 90\% on 1.6 million people. We show that recognition rates on unconstrained data are much lower. \cite{ortiz2014face} experiment with large scale identification assuming there is more than one input photo per person. Recently, several papers reported identification results  on the LFW dataset~\cite{best2014unconstrained, taigman2014web, sun2014deeply, sun2015deepid3}.

\section{The 1 Million Faces Dataset}

\begin{figure*}
	\begin{center}
		\includegraphics[width=0.8\linewidth]{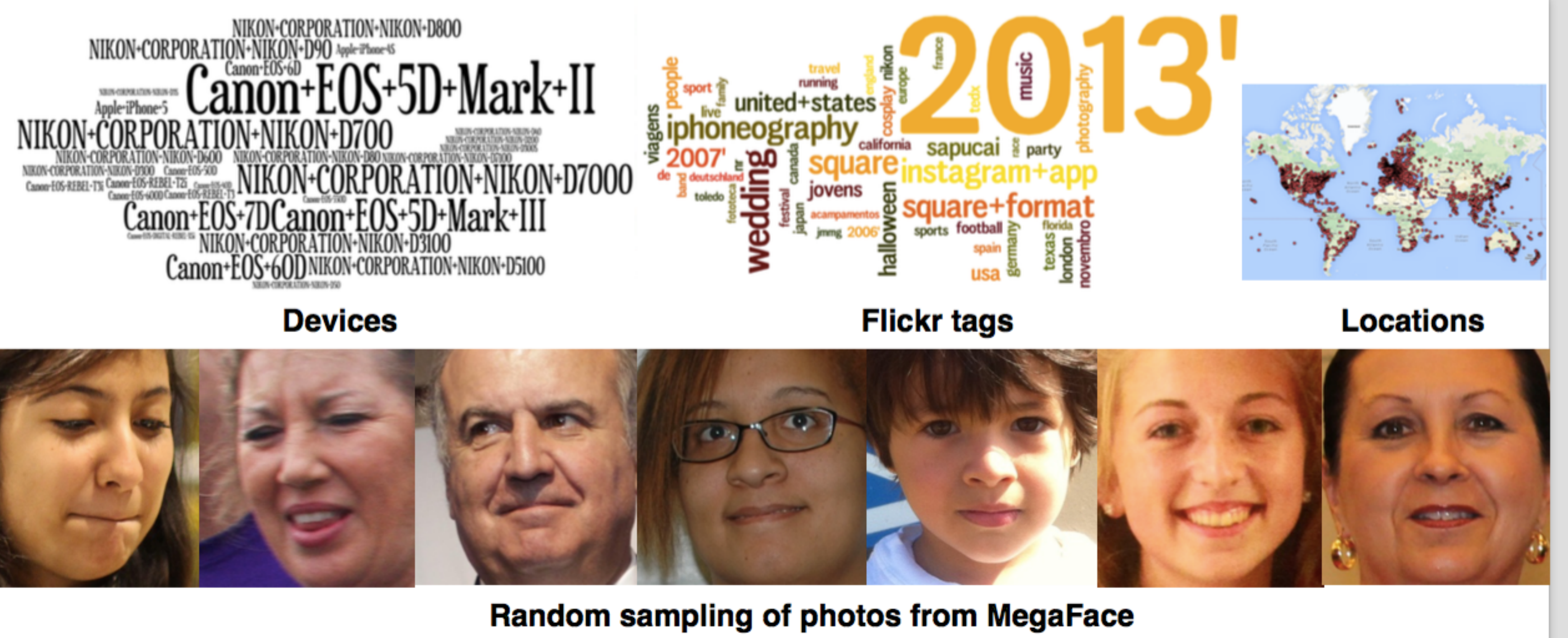}
	\end{center}
	\caption{The MegaFace dataset: distributions of devices, Flickr tags,  and location. We also show a random sample of the photos in the dataset. All the 1 Million photos in the dataset are creative commons photos and will be released for research.}
	\label{fig:data_stats} 
\end{figure*}

\begin{figure}
	\includegraphics[width=.49\linewidth]{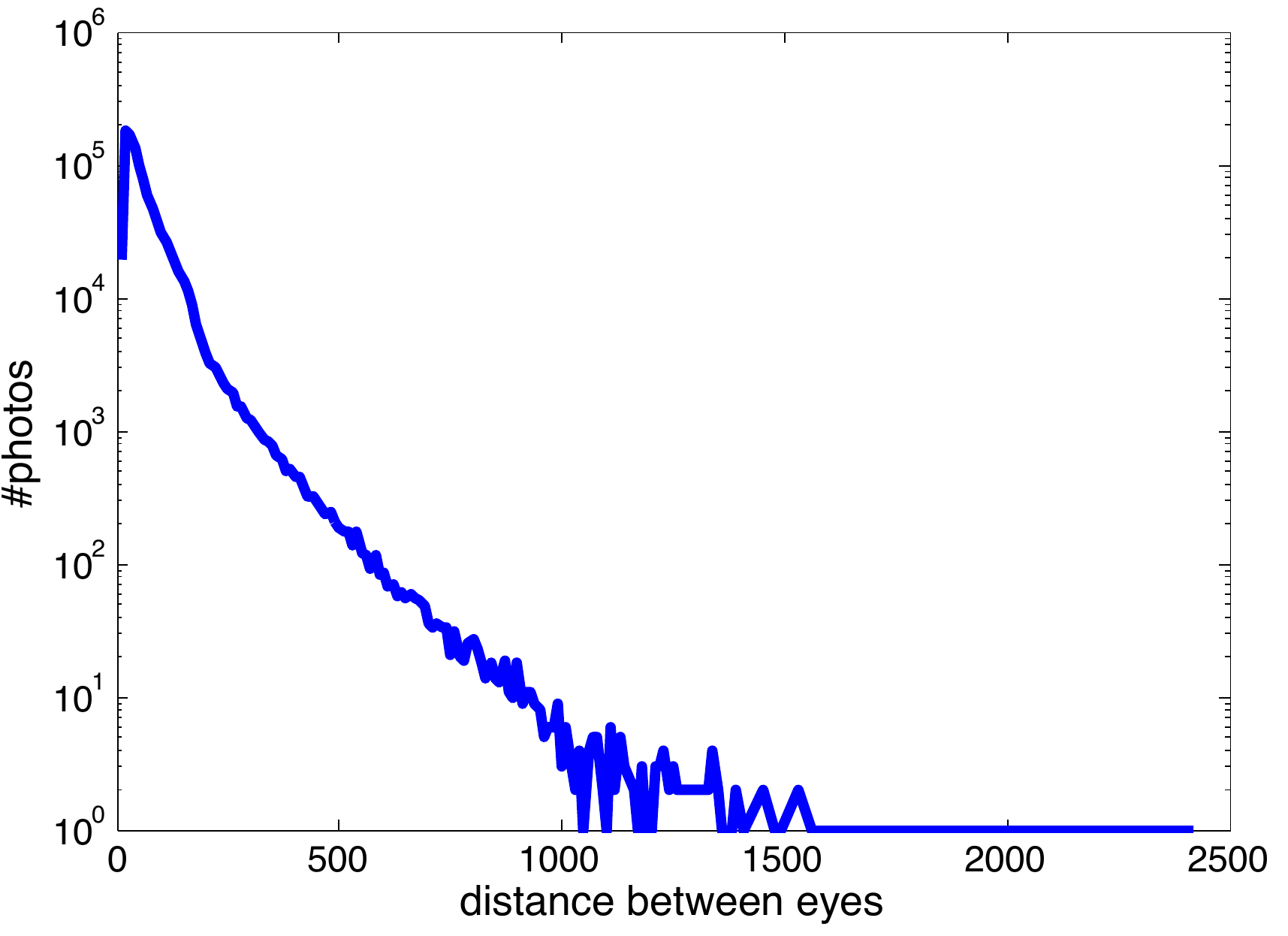}
	\includegraphics[width=.49\linewidth]{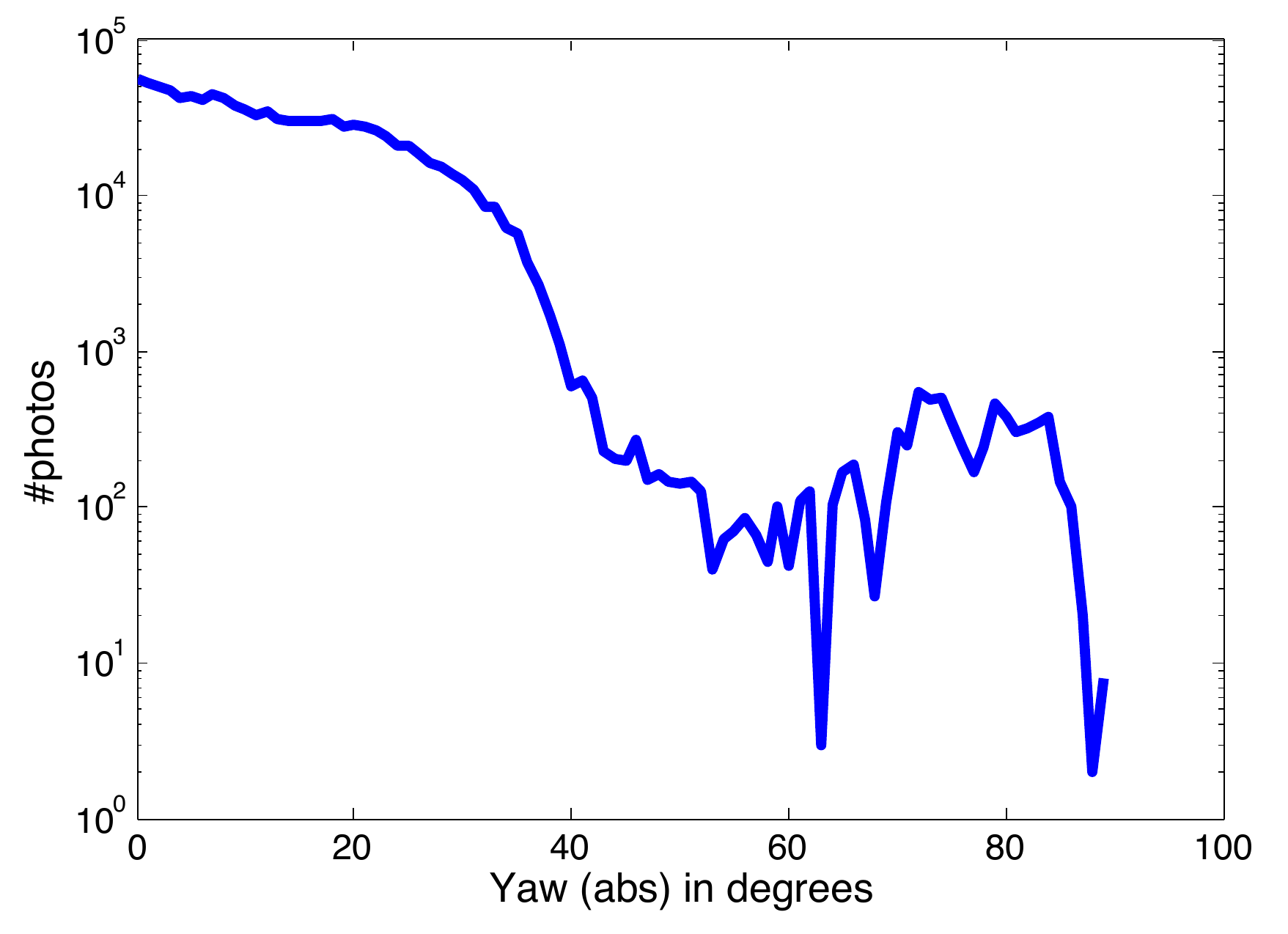}
	\caption{Number of photos per resolution (left) and distribution of poses in MegaFace. Resolution is measured by the distance between the eyes (x-axis). For comparison LFW's distance is 40 while our MegaFace dataset has a wider distribution since Flickr photos are typically high resolution photos taken with DSLR cameras. }
\end{figure}

Our goal is to evaluate how the size of a dataset affects recognition rates.
Specifically we're interested in varying the number of ``distractors'', i.e., people that are not in the test set, and evaluating performance of current face recognition algorithms. 
For example, consider the scenario of identifying a  person from a single photo, by comparing  that image (the "probe") to a database of billions of other people.
This would require several orders of magnitude more comparisons than LFW, which only involves a few thousand people.
We decided that a good starting point would be to evaluate recognition with a million people.
Specifically, we had the following challenges: 
\begin{itemize}
	\item unconstrained in-the-wild photos, i.e., any lighting, pose, expression,  age, and resolution 
	\item photos of regular people, i.e., not  celebrities 
	\item 1 million photos 
	\item publicly available to enable further  experiments by other   researchers 
	\item broad rather than deep dataset, i.e., large number of identities versus  many photos of a small set of people. 
\end{itemize}
While there are massive amounts of photos on the Internet, satisfying the above requirements turned out to be not trivial. Indeed, previous works that released public datasets focused  on photos of celebrities or constrained environments, and collected photos of at most 10K different people. Governmental datasets, e.g., drivers licenses, mugshots, etc. are not available for research and typically captured under constrained conditions. Datasets that are collected by Google, Facebook, Face++ and others, similarly are not available for research. 

This year, however, Yahoo  released their 100 Million Flickr creative commons dataset \cite{thomee2015new}, which turned out to be a perfect opportunity to test face recognition at scale.
We randomly sampled images from this dataset, detecting faces in the photos until we collected a million faces.
These photos are uploaded by Flickr users under the creative commons license and thus the chance of celebrities occurring is exceedingly small (photos of celebrities are typically not available under the creative commons license). We do not guarantee 1 million unique faces (different people) however we optimized for  unique users  and large fraction are group photos which ensure   that the people in the photo are different. Specifically, we have downloaded photos from $178,452$ different users. Given that people appearing in the same group photo are different, we assembled total of $690,572$ unique faces (counting a single photo per user). The rest of the faces may be unique as well, however, we cannot guarantee this. This is the largest set of unique people ever assembled.

\textbf{Data collection protocol.} We began by estimating that there are 500K userids in the Flickr set. Our algorithm for detecting and downloading faces was as follows. For each userid we consider previously unseen photo (based on name, quality, and resolution) and test whether it is possible to detect faces in this photo. If the photo does not have faces we continue to the next one in that user's set, otherwise we detect all the faces in the photo (many of the photos are group photos). Once a photo with faces is found for a particular user we continue to the next user. We terminate the process once we've reached a million faces.  We downloaded the highest resolution available per photo. The faces are detected using the HeadHunter\footnote{\url{http://markusmathias.bitbucket.org/2014_eccv_face_detection/}} algorithm by Mathias et al. \cite{Mathias2014Eccv}, which reported state of the art results in face detection (see exact detection rates and comparisons to others methods in their paper).  We also found that it is robust to extreme face poses.   For each face we save  the detected face (with the face taking about 75\% of the photo). We further estimate  49 fiducial points, and yaw and pitch angles, all calculated by the IntraFace\footnote{\url{http://www.humansensing.cs.cmu.edu/intraface/}} landmark model \cite{xiong2013supervised}.

\textbf{Data statistics.} In Figure~\ref{fig:data_stats}, we present statistics of the dataset. Specifically, we plot distributions of resolution, location, pose, devices, tags, and user ids. The dataset has good distribution of locations, most of the photos were captured by DSLR cameras, tags include words from 'instagram' to 'wedding' which suggests a range of photos from selfies to high quality portraits (large amount of the photos came with a tag '2013' due to nature of the dataset--recently uploaded photos). We also report the distribution of yaw angle and resolution of the face via the distance in pixels between the eyes.  More than 50\% (514K) of the  photos in MegaFace have resolution more than 40 pixels  inter-ocular distance (which is the resolution of LFW photos). In addition, we inspected by running stricter face detectors on the automatically downloaded photos, as well as, manually inspected 1500 random photos,  and found that 6\% of the photos are very blurry, non faces, or very low resolution.

%

\section{Evaluation Methodology}

Given the dataset, we ran several experiments to examine two classic recognition scenarios: identification and verification.
In addition, we ran a large scale human performance evaluation on the identification task.
Below we describe the methodology and our test set. 

\paragraph{Test set.} Our Flickr dataset is used to create a large number of distractors. For testing known identities we use a subset of the FaceScrub dataset that was created in 2014 by Ng and Winkler \cite{ng265data}.
FaceScrub includes 100K photos of 530 celebrities.  It  can be freely downloaded from {\footnotesize \url{http://vintage.winklerbros.net/facescrub.html}}. 
We chose to use this dataset as our test set (rather than LFW) since it has a similar number of male and female photos (55,742 photos of 265 males and 52,076 photos of 265 females) and a large variation across photos of the same individual. Ensuring variation within individual's photo collection is important to remove possible bias, e.g.,  due to backgrounds and hair style \cite{kumar2011describable}, that may occur in LFW. We randomly selected a subset of FaceScrub to create a set which is comparable to LFW in size but has more variation across identities. We randomly selected 80 identities with that had more than 50 images each, for a total of 4000 faces.

\paragraph{Identification and Verification.} Identification means that given a photo of a person as input and assuming there is a photo of this person in the database together with many other people, the algorithm should match the two photos of the same person. Verification means that given a pair of photos, the algorithm should output whether the person in the two photos is the same or not. 
Until now mostly verification was at the focus of face recognition research, and tested by the LFW benchmark \cite{huang2007labeled}. Recently, \cite{best2014unconstrained, taigman2014web, sun2014deeply, sun2015deepid3} performed also identification experiments on LFW. In this paper, we explore both scenarios but with very large number of  distractors.  

Specifically, each person was enrolled to the database with a single photo, i.e., the database included a million Flickr photos and photo per test identity.
Then for each person we used as test photo every photo from the collection except for the database one.
We then averaged the results over the different photos.
We report the results as Cummulative Match Characteristics curves.
This curve shows the probability that the correct gallery image will be chosen for a random probe by rank-K in a list of results.
To evaluate verification we computed all pairs between the probe database (FaceScrub) and the distractor database (Flickr).
This means that our verification eperiments has in total 4 billion negative pairs.
We report verification results with ROC curves; this explores the tradeoff between falsely accepting non-match pairs and falsely rejecting match pairs.

\paragraph{Recognition Methods.}
We have selected to experiment with four recognition algorithms that represent very different types of techniques. 
\begin{itemize}
	\item \textbf{Basic LBP comparison.} We have implemented a  comparison based on Local Binary Pattern (LBP) descriptors \cite{ahonen2006face}.  This approach achieves 70\% recognition rates on LFW, and uses no training. 
	\item  \textbf{Joint Bayesian.}   The Joint Bayesian model represents each face as the sum of two Gaussian variables
$x = \mu + \epsilon$ where $\mu$ represents identity and $\epsilon$ represents inter-personal variation.  To determine whether two faces, $x_1$ and $x_2$ belong to the same identity, we calculate $P(x_1, x_2 | H_1)$ and $P(x_1, x_2 | H_2)$ where $H_1$ is the hypothesis that the two faces are the same and $H_2$ is the hypothesis that the two faces are different. These distributions can also be written as normal distributions, which allows for efficent inference via a log-likelihood test. This algorithm, trained on \cite{yi2014learning} achieves 89\% on LFW.

\item \textbf{Commercial software by VisionLabs.} 
VisionLabs  \small{\url{http://www.visionlabs.ru/}} achieved 93\% recognition rates on LFW, and is trained on outside data. 

\item \textbf{Google's FaceNet.} Google FaceNet \cite{schroff2015facenet} is the most recent and highest performing of several Deep Learning algorithms applied to the LFW benchmark. 
Unlike DeepFace and DeepID  which have a bottleneck layer and are optimized by minimizing cross-entropy, FaceNet learns an embedding such that the extracted features are directly comparable using the Euclidean distance.
It is  trained on 200  Million photos of 8 Million people, and achieves 99.6\% on LFW. 
\end{itemize}

\section{Human Performance}
While a lot of effort goes into developing automated face recognition algorithms, still the human visual system seems to perform better especially at very low false accept rate. 
It is therefore very interesting to estimate the human recognition rate on the same sets of photos on which algorithms operate. 
Verification  rates of humans   were previously estimated  on controlled data, i.e., photos taken in laboratory condition \cite{o2007face, o2012comparing, phillips2010frvt,adler2007comparing}, and more recently on unconstrained photo collections: \cite{kumar2011describable} evaluated verification rates on the LFW dataset, and \cite{best2014unconstrained, chen2012dictionary} estimated verification rates on videos. Human studies made on unconstrained photos, e.g., \cite{kumar2011describable}, fused human judgments by averaging ratings over participants, which helped  remove outliers. 
Until recently, none of the algorithms was able to outperform humans on LFW. Moreover, all the previous human experiments were done on small scale and did not evaluate identification. It is of great interest, however, to discover how humans perform at \textit{scale} to provide a lower bound for machine performance.


One of the key contributions of this paper is an extremely large (about 4 million pairs of faces) human study on Mechanical Turk that evaluates human performance on unconstrained photos (Flickr), and specifically targets identification rather than verification. 

We have performed the following experiment on Mechanical Turk.  Since all the identities in the FaceScrub dataset (our test set) are celebrities, human recognition rates may be biased due to familiarity with the person \cite{sinha2006face}. We therefore sorted all the names in FaceScrub according to the number of results that Google image search returns per person as a measure of popularity. We then chose 50 most popular people, and 50 least popular people as our human experiment test set. Each person had 100 photos, we randomly selected one photo as the probe image, and used the rest 99 as gallery images. We then produced 99 positive pairs per person.   For the distractor set, for each input photo we randomly selected 10K photos from our MegaFace dataset, and produced 10K pairs of probe with each of the distractors. This results in total of $100\times (99+10K)$ pairs. Since the number of positive pairs in this setting is very low, we introduced additional positive pairs by randomly pairing gallery images that are not the probe. This is to remove possible bias in human rating, i.e., if most pairs are negative people may miss the positive ones. We presented to turkers 10 pairs per page and asked to click on all the pairs that contained the same person. We paid 1 cent per page of 10 pairs.

Once this experiment was done we collected the pairs that received 1 click or more, and created a sorting experiment.
We selected only the pairs that include the probe photos, and created a set of possible matches per probe.
We generated triples of probe, and two matches, presented 10 triples in each page and asked which one of the matches is the person in the probe. Generally, to get a full ranking of all images, the number of possible triples per probe is $n^2$ where $n$ is the number of matches from round 1. For efficiency (and less cost) only determined the position of each gallery photo relative to the distractor images. That is, our experiment determined the number of distractors that would be ranked above and below each gallery image, but not the ordering within those groups. On every pair/triple of photos in both experiments worked 3 different people. We paid 7 cents for each page of 10 triples. The total cost of this experiment was \$10,000.



\begin{figure}
	\begin{center}
   \caption*{Identification}
  \begin{tabular}{ c | c | c }
    \hline
    & Rank-1 & Rank-10 \\ \hline
    All & 23.9 & 91.13 \\ \hline
    Males & 23.35 & 89.98 \\ \hline
    Females & 24.01 & 92.5 \\ \hline
    Less Popular & 22.7  & 90.9 \\ \hline
    More Popular & 25.1 &  91.3 \\
    \hline
  \end{tabular}
\end{center}

  \begin{center}
  \caption*{Verification}
  \begin{tabular}{ c | c | c }
    \hline
    & TAR @ $2 \times 10^{-3}$ & TAR @ $5 \times 10^{-2}$ \\ \hline
    All & 41.6 & 76.5 \\ \hline
    Males & 43.7 &  79.0 \\ \hline
    Females & 39.4 & 73.9  \\ \hline
    Less Popular & 39.4 & 74.7 \\ \hline
    More Popular & 43.6 & 78.2 \\
    \hline
  \end{tabular}
\end{center}

	\caption{Human recognition rates (verification and identification).
  Our experiments also show that humans perform  better on more popular people and are  better at the verification task when comparing males.}
	\label{fig:human_id}
\end{figure}

	\section{Benchmarking Recognition at Scale}

This section describes the verification and  identification experiments   we performed with  the MegaFace dataset as our distractor set.  In addition, we describe the effect of pose variation on recognition at scale.  We have experimented with four automatic face recognition algorithms: LBP and Joint Bayes were implemented by us, VisionLabs has provided their software for our experiments, and FaceNet algorithm was ran by the authors on our data.  Prior to our experiments, we have verified for all methods that they  achieve their reported results \cite{huang2007labeled} on the LFW dataset.  Similarly, we repeated the human study of \cite{kumar2011describable} on LFW using our Mechanical Turk interface to ensure that our results are valid (Fig. 2  in the sup. material).

		\begin{figure*}
			\centering
			\includegraphics[width=.4\linewidth]{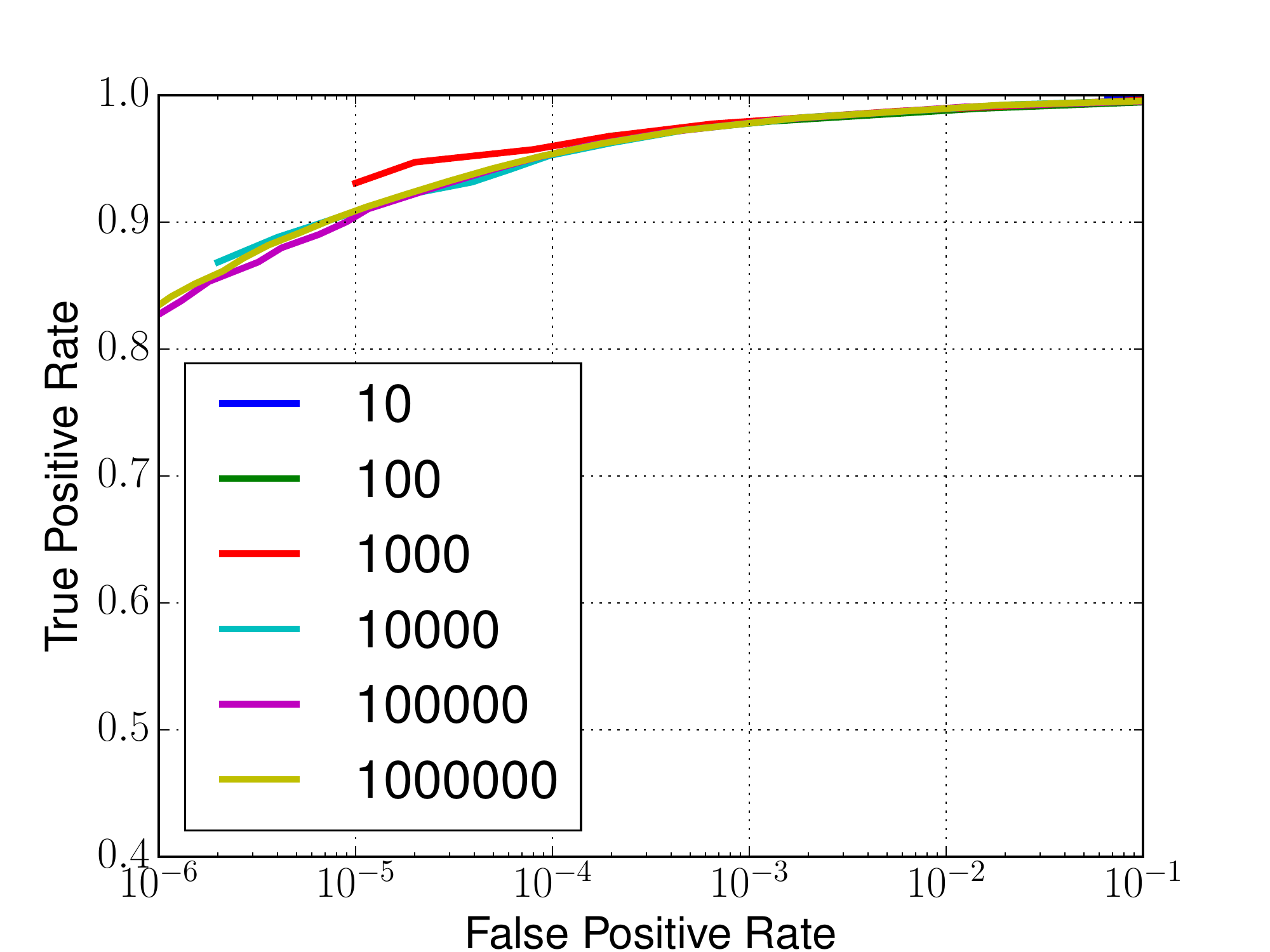}
			\includegraphics[width=.4\linewidth]{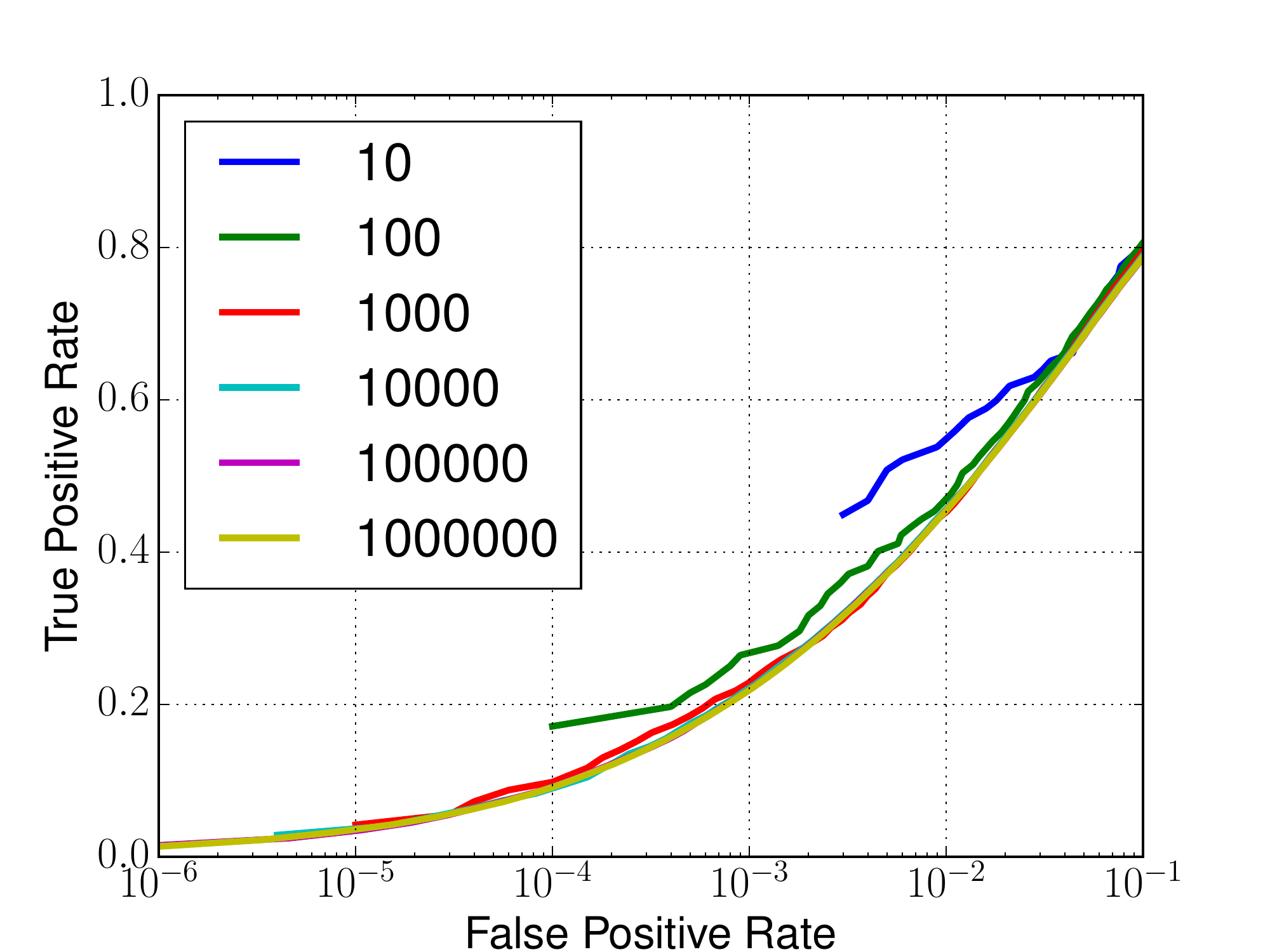}
			\caption{Verification performance with 10, 100, 1K, 10K, 100K, and 1 Million distractors, for two methods: FaceNet (left) and Joint Bayesian (right). Verification rates are stable with different database sizes. }\label{fig:verf}
		\end{figure*}
	
	\begin{figure*}
		\begin{center}
			\includegraphics[width=0.4\linewidth]{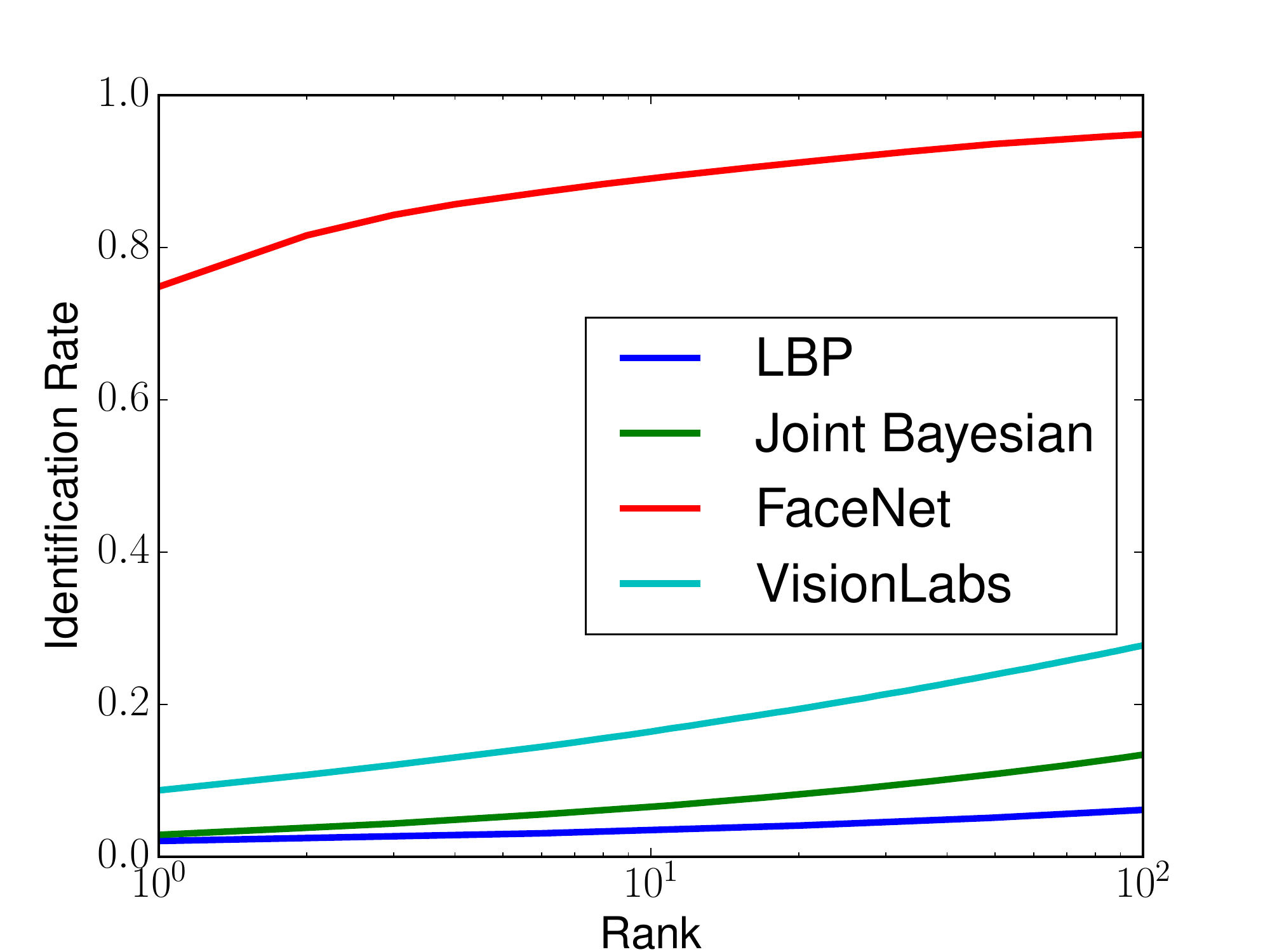}
			\includegraphics[width=0.4\linewidth]{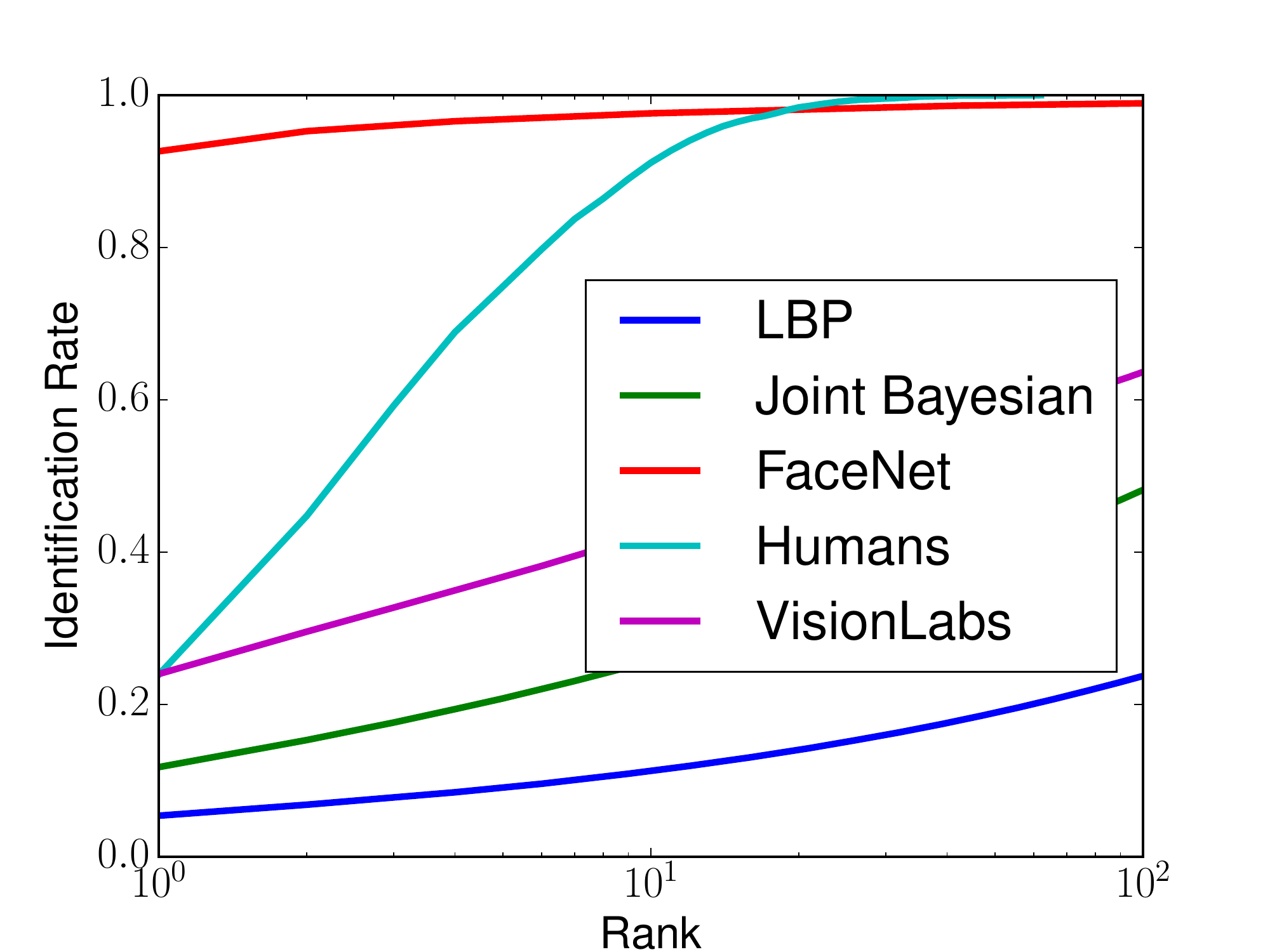}
		\end{center}
		\caption{Identification performance for all methods with 1M distractors (left) in the database, and 10K distractors (right) which also includes human performance. }
		\label{fig:dataset_size_cmc}
	\end{figure*}

		\paragraph{Verification.} Fig.~\ref{fig:verf} shows results of a verification experiment of two algorithms (FaceNet and Joint Bayesian) with various numbers of distractors going from 10 to 1 million. We make two observations: 
		
		\begin{itemize}
			\item  Verification does not change much at scale, particularly when we consider false accept rates as in LFW. Results at LFW are typically reported at equal error rate which implies false accept rate of 1\%-5\% for top algorithms. We believe the reason that rates stay constant at scale is because given a probe photo, if the face has 100 other faces that can be matched wrongly in a small dataset, e.g., thousand faces, assuming uniform distribution of the data, the rate will stay the same, and so in a dataset of a million faces one can expect to find 10,000 matches at the same false accept rate.  
			\item Striving to perform well at low false accept rate  is important with large datasets. Even though the chance of a false accept on the small benchmark is acceptable, it does not scale to even moderately sized galleries.
		\end{itemize}

		\paragraph{Identification.} 
		
	The relation between the identification and verification protocols was studied by Grother and Phillips \cite{grother2004models} and DeCann and Ross \cite{decann2012can}.  Only recently, however,  identification results have started to appear on the LFW dataset. Here we evaluate identification with large number of distractors. In Figs.~\ref{fig:teaser} and \ref{fig:dataset_size_cmc} we show the performance of the four algorithms with respect to different ranks, i.e., rank-1 means that the correct match got the best score from the whole database, rank-10 that the correct match is in the first 10 matches, etc. Fig~\ref{fig:teaser} shows that rates drop significantly at scale for everyone that we tested except for FaceNet which exhibits a relatively small decrease. Fig.~\ref{fig:dataset_size_cmc} shows the behavior at higher ranks.

		\paragraph{Pose.}

		In general, frontal faces are  considered to be  easier to match since features are directly comparable and alignment is well understood.
		However, when we removed all non-frontal images from our Flickr dataset and repeated our experiment, the reuslts were surprising: rates  dropped (Fig.~\ref{fig:frontal}). 
		 We have created a distractor set of only frontal photos (yaw$<$2 degrees), and only non-frontal (yaw$>$15 degrees). We tested the Joint Bayesian, and FaceNet algorithm with this set. We got that identification rates are lower by about 5\% in case of a frontal distractor set for the Join Bayesian algorithm.  FaceNet that performs better than JB on the full set, does not exhibit significant difference between frontal and non frontal distractors. 
		 
		 A number of factors could explain this.
		Our primary hypothesis is that our probe dataset is biased towards frontal images, and are thus less likely to match to non-frontal images in the distractor database.
		Another plausible explanation of this effect is that algorithms with limited learning capacity must trade performance for a given pose for generalization across pose; i.e., an algorithm that is able to match across pose might perform worse at frontal recognition.
		
\begin{figure*}
				\centering
				\includegraphics[width=0.4\linewidth]{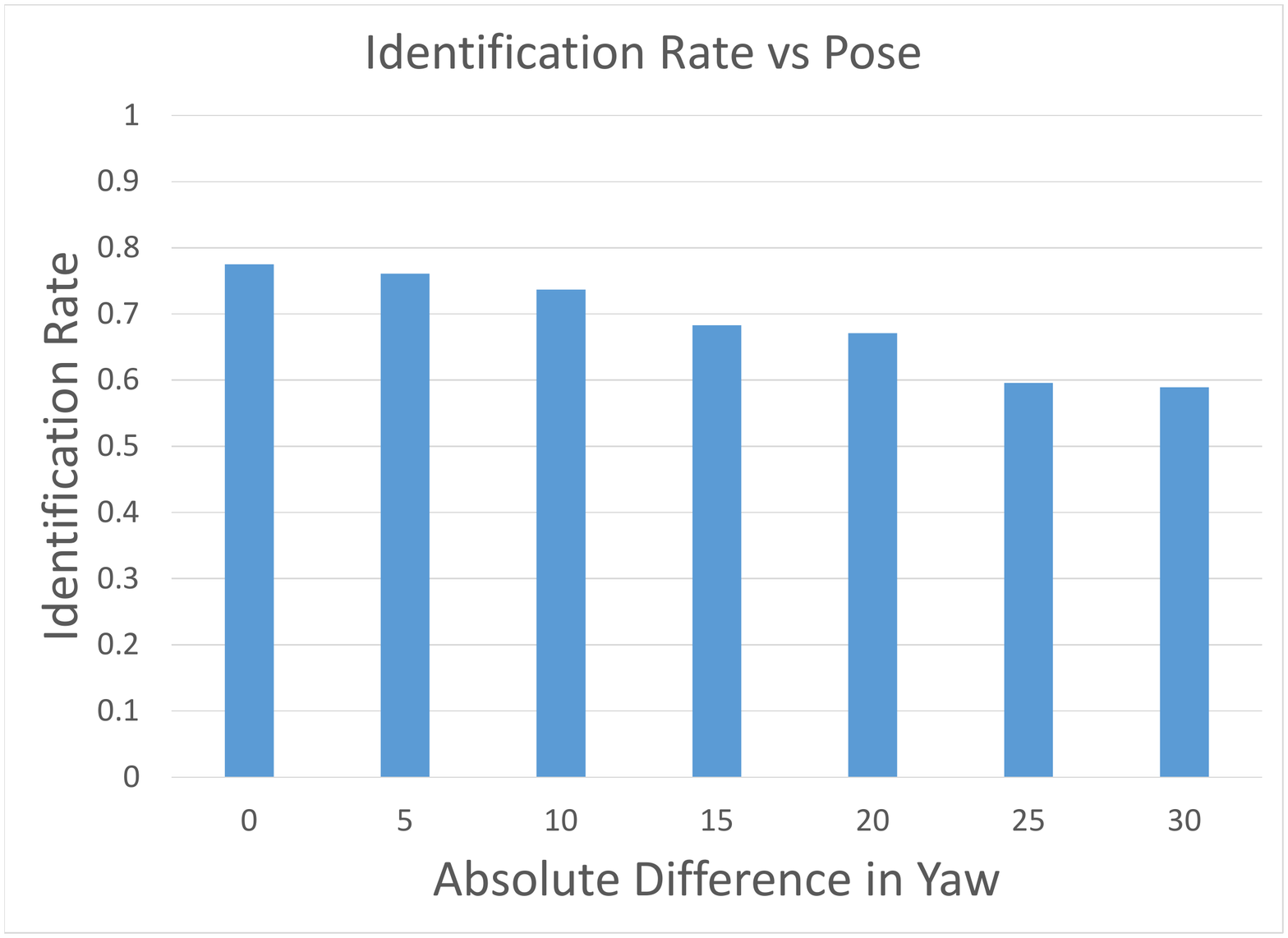}
				\includegraphics[width=0.4\linewidth]{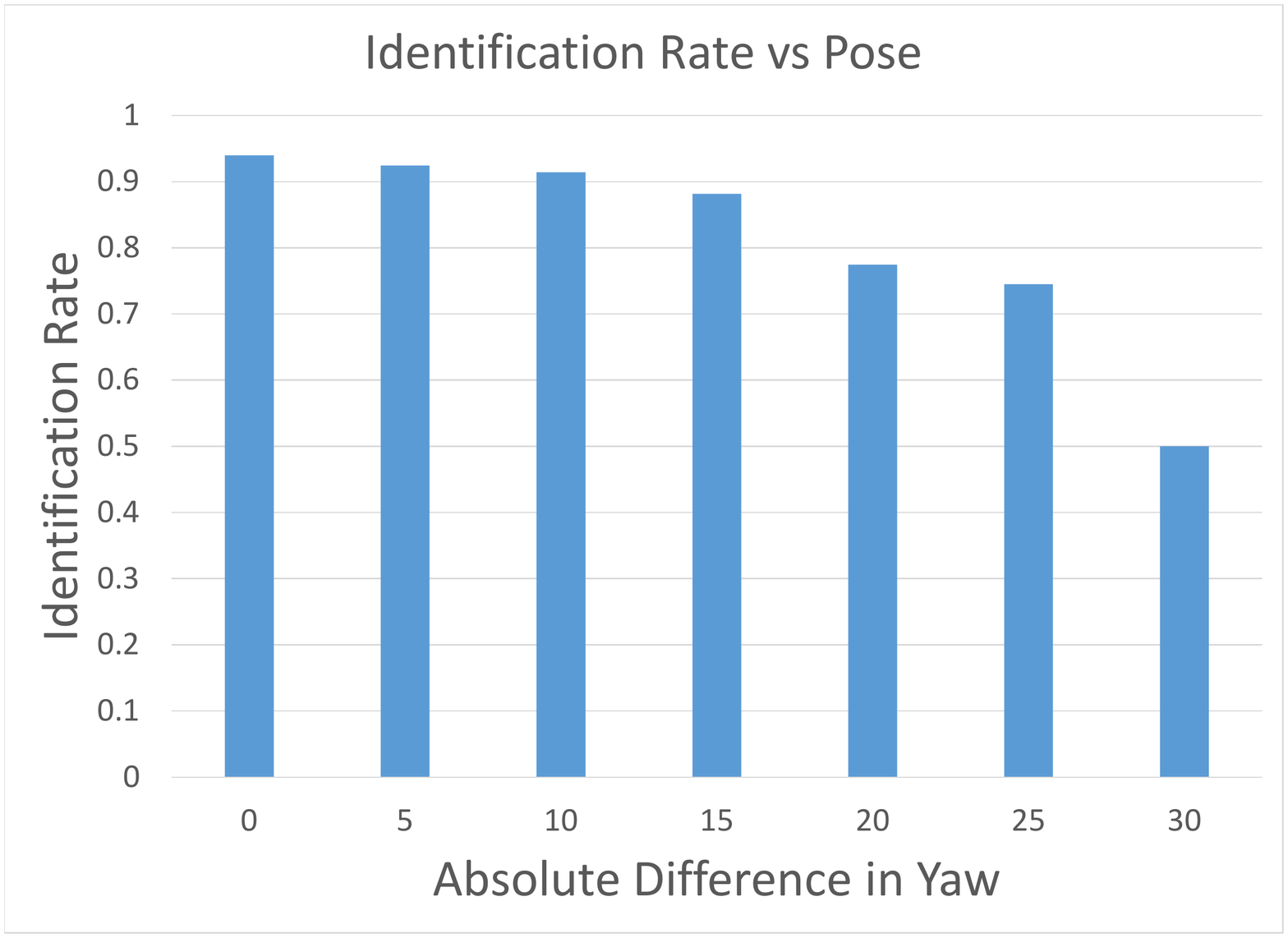}
				\caption{Identification performance of FaceNet (left plot) and humans (right plot) as a function of pose. FaceNet (left):  we can see that performance decreases with pose difference (between the probe and gallery) even with the best method.   This indicates that cross-pose matching is still an open problem. Humans (right): similarly to automatic methods humans perform worse with difference in pose.}\label{fig:googlepose}
			\end{figure*}

\begin{figure}
				\includegraphics[width=.9\linewidth]{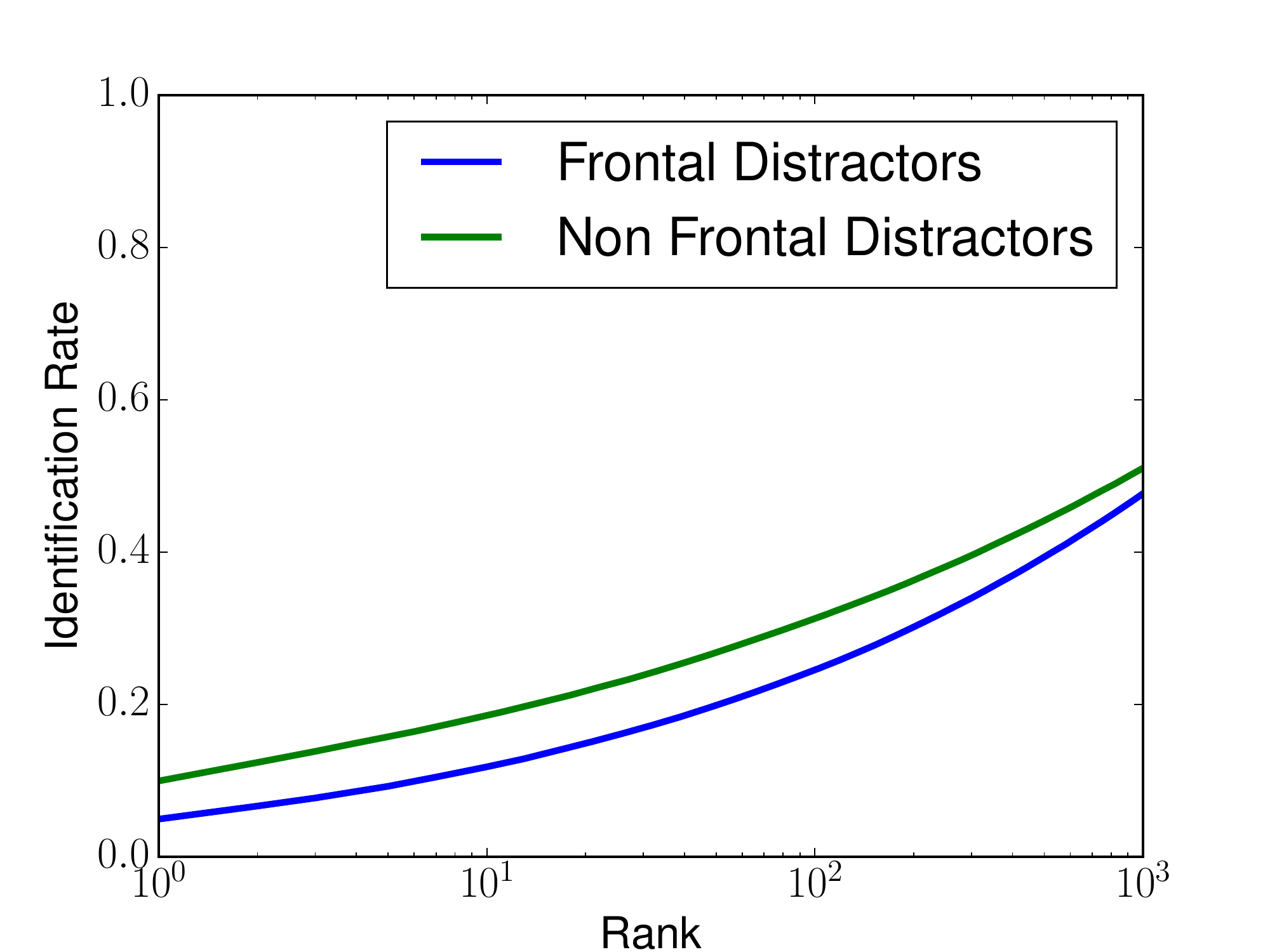}
				\caption{Cumulative Match Characteristics of the Joint Bayesian method with 100K distractors in the database. Recognition rates are lower in case the distractors appear in frontal pose (yaw $<2$ degrees).  }\label{fig:frontal}
			\end{figure}

		We have also experimented with difference in pose within the pair of photos. Fig.~\ref{fig:googlepose} evaluates error in recognition with respect to difference in yaw between the probe and gallery. We can see that larger pose difference implies larger errors. The results are normalized by the total number of pairs for each pose difference.

\section{Discussion}

Ultimately, a face recognition algorithm should be able to perform even with billions of people in the dataset. While testing with billions is still challenging we have done the first step and created a database of a million faces. This dataset will be available to researchers and we hope that more methods will be tested and improved using the provided data. 

In the future, we  intend to release all the detected faces from the 100M Flickr dataset. While companies like Google and Facebook have a  head start due to availability of enormous amounts of data, we're interested to provide an even playing field for evaluation and training of algorithms at scale. A big challenge is, can one come up with high identification rates by training on our Flickr data. Our dataset will be separated to testing and training sets for fair evaluation and training. This playing field point is particularly important, as the Google + Facebook methods are trained on orders of magnitude more imagery than the others.

We will maintain and update this benchmark online and solicit contributions from the other top performers (e.g., we are in contact with other companies and hope to include additional results before the paper goes out to press). Our first challenge is to test identification and verification with 1 million distractors, while the FaceScrub dataset is used as the test set.  We plan to keep creating more challenges in the future. One challenge will be to create a test set from Flickr photos. An interesting question is whether results will change if test set does not include celebrity photos but regular Flickr photos.  Of course dataset bias will be taken into account. We will also test different sizes of test set going from 80 identities (current challenge) to hundreds of thousands. Another challenge will be to allow  multiple photos rather than a single photo per person for identification. Finally,   the significant number of high resolution faces in our Flickr database will also allow  to explore the issue of resolution in more depth. Resolution is mostly untouched topic in face recognition literature, mostly because public data was not available.

{\small
\bibliographystyle{ieee}
\bibliography{paper_minor_revision}
}

\newpage
\section{Supplementary material}

\begin{figure}[h]
	\begin{center}
		\includegraphics[width=1\linewidth]{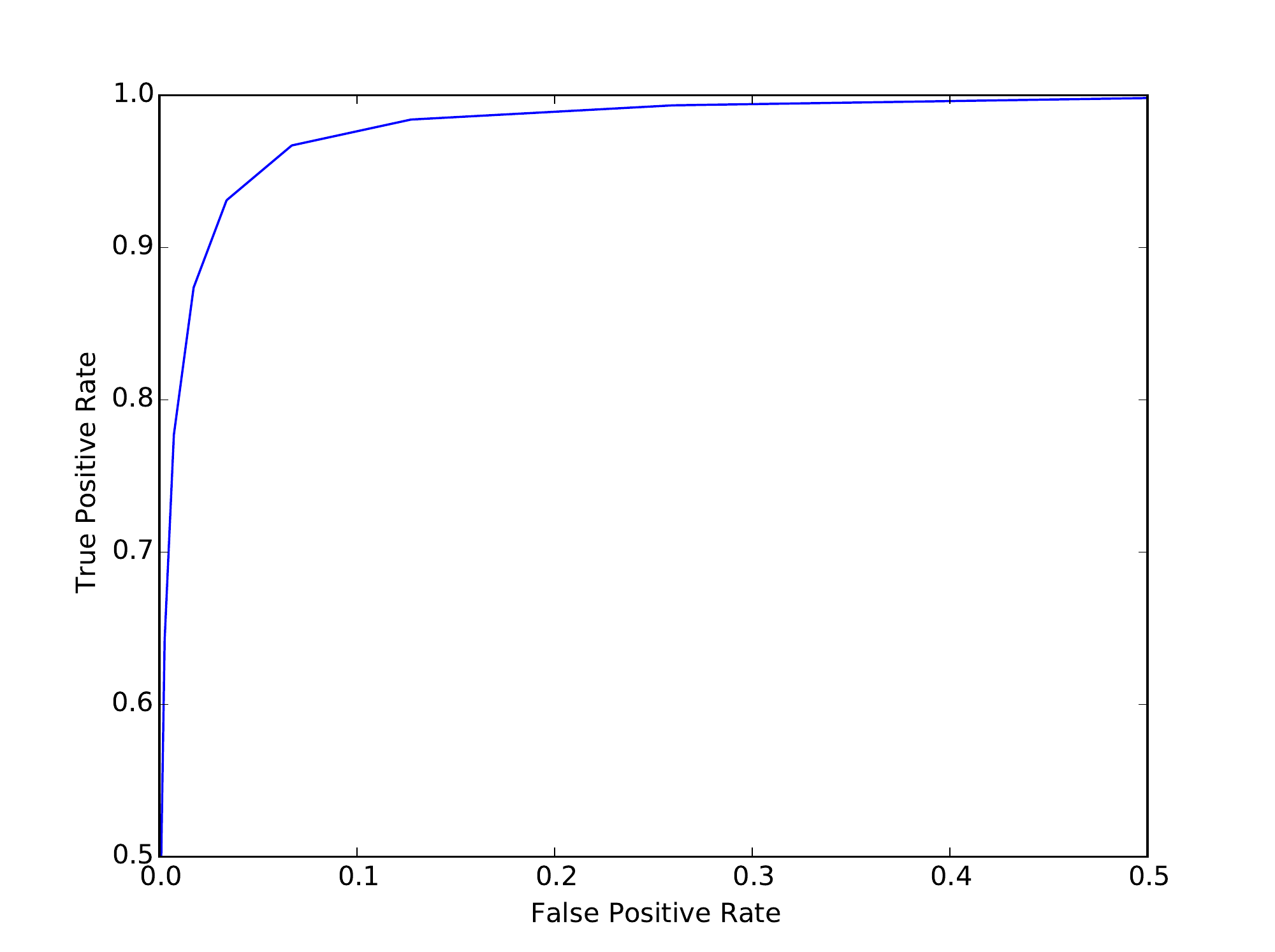}
	\end{center}
	\caption{In order to verify that our interface did not affect the performance of Human Recognition, we repeated the LFW experiment by Kumar et al. with our interface -- that is, for each of the 6000 pairs in LFW we asked 10 workers to decide whether each pair was same / not same. The rates are slightly lower than Kumar's et al. results (0.9576 vs 0.9753), but still very good. We suspect that the difference is since we asked only same/not same question (binary) while Kumar et al. asked to rank on a scale of 0 to 5. }
\end{figure}

\begin{figure}[h]
	\begin{center}
		\includegraphics[width=1.0\linewidth]{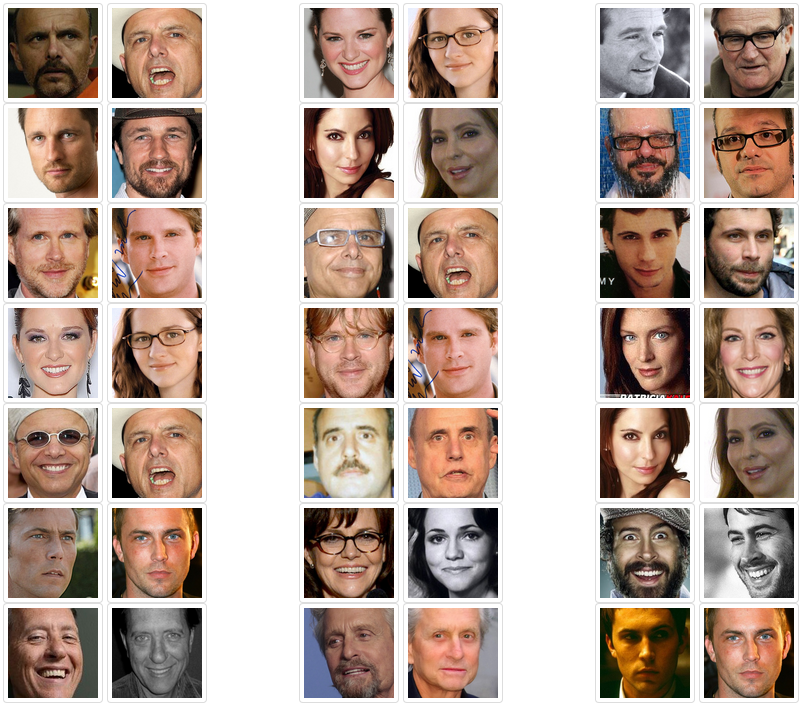}
	\end{center}
	\caption{Random pairs of probe/gallery images that were matched correctly by FaceNet, but not by Humans. \href{HumanMissFaceNetRight.html}{Click here to view the full set of faces.}}
\end{figure}

\begin{figure}[h]
	\begin{center}
		\includegraphics[width=1.0\linewidth]{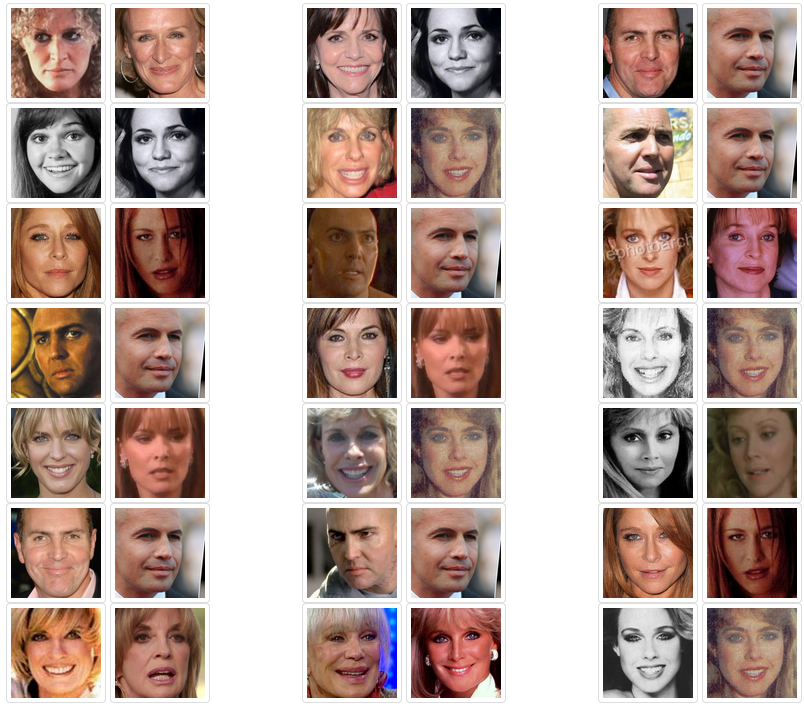}
	\end{center}
	\caption{Random pairs of probe/gallery images that were matched correctly by Humans, but not by FaceNet. \href{FaceNetMissHumansRight.html}{Click here to view the full set of faces.} }
\end{figure}

\begin{figure}[h]
	\begin{center}
		\includegraphics[width=1.0\linewidth]{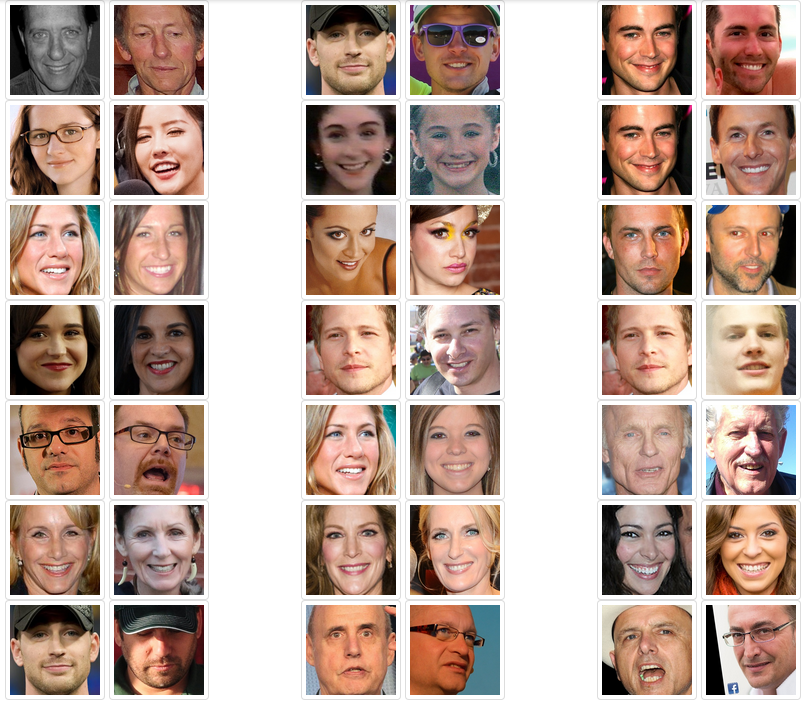}
	\end{center}
	\caption{Faces incorrectly identified as being the same by Humans. \href{HumanFalsePositives.html}{Click here to view the full set of faces.}}
\end{figure}


\end{document}